\documentclass[letterpaper]{article} 

\ifdefined\aaaianonymous
    \usepackage[submission]{aaai2026}  
\else
    \usepackage{aaai2026}              
\fi

\usepackage{times}  
\usepackage{helvet}  
\usepackage{courier}  
\usepackage[hyphens]{url}  
\usepackage{graphicx} 
\usepackage{natbib}  
\usepackage{caption} 
\usepackage{algorithm}
\usepackage{algorithmic}

\usepackage{newfloat}
\usepackage{listings}
\DeclareCaptionStyle{ruled}{labelfont=normalfont,labelsep=colon,strut=off} 
\floatstyle{ruled}
\newfloat{listing}{tb}{lst}{}
\floatname{listing}{Listing}

\usepackage{amsmath}
\usepackage{amssymb}
\usepackage{bm}

\usepackage{multirow}
\usepackage{array}
\usepackage{xcolor}

\title{MolSight: Optical Chemical Structure Recognition with SMILES Pretraining, Multi-Granularity Learning and Reinforcement Learning}

\author{
    Wenrui Zhang \quad
    Xinggang Wang \quad
    Bin Feng \quad
    Wenyu Liu\thanks{Corresponding author: \protect\url{liuwy@hust.edu.cn}}
}
\affiliations{
    School of Electronic Information and Communications, Huazhong University of Science and Technology\\

}

\begin{document}

\maketitle

\begin{abstract}
Optical Chemical Structure Recognition (OCSR) plays a pivotal role in modern chemical informatics, enabling the automated conversion of chemical structure images from scientific literature, patents, and educational materials into machine-readable molecular representations. This capability is essential for large-scale chemical data mining, drug discovery pipelines, and Large Language Model (LLM) applications in related domains. However, existing OCSR systems face significant challenges in accurately recognizing stereochemical information due to the subtle visual cues that distinguish stereoisomers, such as wedge and dash bonds, ring conformations, and spatial arrangements.
To address these challenges, we propose \textbf{MolSight}, a comprehensive learning framework for OCSR that employs a three-stage training paradigm. In the first stage, we conduct pre-training on large-scale but noisy datasets to endow the model with fundamental perception capabilities for chemical structure images. In the second stage, we perform multi-granularity  fine-tuning using datasets with richer supervisory signals, systematically exploring how auxiliary tasks—specifically chemical bond classification and atom localization—contribute to molecular formula recognition. Finally, we employ reinforcement learning for post-training optimization and introduce a novel stereochemical structure dataset. Remarkably, we find that even with MolSight's relatively compact parameter size, the Group Relative Policy Optimization (GRPO) algorithm can further enhance the model's performance on stereomolecular. Through extensive experiments across diverse datasets, our results demonstrate that MolSight achieves state-of-the-art performance in (stereo)chemical optical structure recognition.
\end{abstract}

\ifdefined\aaaianonymous
\else
\begin{links}
    \link{Code}{https://github.com/hustvl/MolSight}
\end{links}
\fi

\section{Introduction}
A vast amount of chemical information is locked within static images in publications, making it inaccessible for computational analysis. Recent document parsing pipelines \cite{wang2024mineruopensourcesolutionprecise, li2025monkeyocrdocumentparsingstructurerecognitionrelation} typically render chemical diagrams as simple images. Optical Chemical Structure Recognition (OCSR) addresses this by converting these images into machine-readable formats. This automated process is crucial for accelerating drug discovery and building chemical databases.


\begin{figure}[t]
    \centering
    \includegraphics[width=0.9\linewidth]{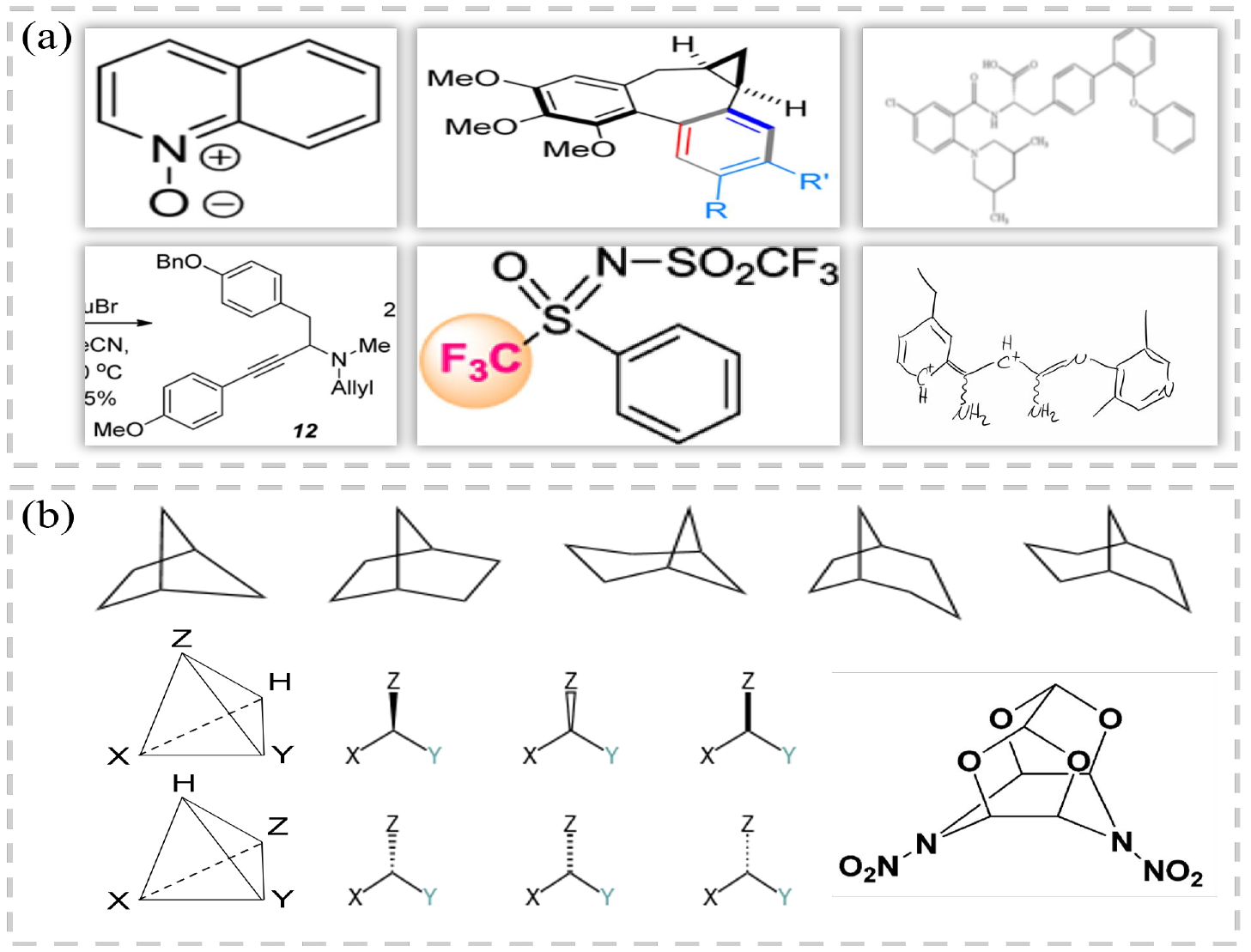}
    \caption{\textbf{Examples of challenging chemical structures images.} (a) Diversity of images from different sources. (b) 3D molecular information encoded within 2D images.}
    \label{fig:cases}
\end{figure}

General OCSR challenges stem from the diversity of chemical structure sources: 
(1) Image Degradation: High variance in quality and artifacts from heterogeneous sources like scans and sketches.
(2) Symbolic Complexity: The need to parse complex chemical notations beyond the core molecular graph.
(3) Stylistic Variance: The lack of standardized drawing conventions, leading to inconsistent representations.

The stereochemical configuration of molecules—includ-ing chirality, geometric isomerism, and conformational details—directly impacts their pharmacological effects and chemical behavior, making precise recognition of these structural features a critical requirement for practical applications. Stereochemical recognition introduces additional layers of complexity. As shown in Fig. 1, critical stereochemical indicators such as wedge bonds (representing bonds projecting toward the viewer) and dashed bonds (indicating bonds extending away from the viewer) require precise detection and classification. 
Furthermore, distinguishing between stereoisomers—molecules with identical connectivity but different spatial arrangements—demands sophisticated understanding of geometric relationships and spatial context that traditional approaches often fail to capture effectively.


To address the aforementioned challenges, we propose MolSight, the first multi-stage learning framework specifically designed for OCSR tasks. In the first stage, MolSight supports pretraining using large quantities of images annotated only with SMILES, with the objective of enhancing the perception capabilities of the image encoder for molecular images. 
In the second stage, by introducing the additional chemical bond head and coordinate head, the performance of the model can be improved during this process. Recently, reinforcement learning (RL) has shown tremendous potential in improving LLMs' ability to solve complex reasoning tasks. 
Inspired by this development, MolSight innovatively introduce RL algorithms into OCSR tasks, utilizing images of stereoisomers that are commonly confused by the model for RL optimization, thereby further enhancing the model's understanding of chemical semantics.

The main contributions of this paper can be summarized as follows:

\begin{itemize}
\item We present MolSight, a comprehensive learning framework for OCSR that enhances model performance across diverse molecular types, particularly stereoisomers, through a three-stage training approach consisting of pre-training, multi-granularity fine-tuning, and RL post-training.
\item MolSight represents the first OCSR system to incorporate reinforcement learning methods. By integrating the GRPO algorithm, the model optimization process overcomes the limitations of token-level accuracy and directly optimizes for chemical semantic correctness, effectively improving recognition accuracy for stereoisomeric molecules.
\item We construct a new annotated molecular image dataset, Stereo-200k, consisting entirely of challenging stereoisomeric molecules that are prone to confusion. This dataset supports MolSight's RL training process and will be made publicly available to the research community.
\item Extensive experiments demonstrate that MolSight achieves state-of-the-art performance in terms of accuracy, similarity, and robustness, outperforming both classical and learning-based methods across most scenarios, while showing broad potential for downstream applications.
\end{itemize}

\section{Related Work}
\vspace{0.3cm}
\subsection{Optical Chemical Structure Recognition (OCSR)}

\subsubsection{Traditional OCSR Methods.}
Early OCSR systems \cite{mcdaniel1992kekule, valko2009clide, filippov2009optical, smolov2011imago, peryea2019molvec} were primarily built on handcrafted rules, relying on predefined logic to map image features to chemical structures. These rule-based systems typically employed traditional image processing techniques—including binarization, denoising, smoothing, and thinning operations—to enhance structural lines and characters. The processed images were then vectorized to convert line elements into vector format, followed by customized optical character recognition (OCR) engines to identify atomic characters. Finally, molecular graph structures were constructed using heuristic methods. Many rule-based OCSR tools continue to receive updates today. 

However, these approaches suffer from limited generalization capabilities and struggle to adapt to variations in drawing styles and image quality. 
Previous studies \cite{clevert2021img2mol} have demonstrated that even minor perturbations to input images can cause substantial drops in recognition accuracy for these methods.

\subsubsection{Deep Learning Approaches.}
End-to-end deep learning methods \cite{clevert2021img2mol, rajan2023decimer, MolScribe} typically employ an encoder-decoder architecture that targets line notations of chemical structures (e.g., SMILES \cite{weininger1988smiles} and InChI \cite{heller2013inchi}) as output. In this framework, image encoders represented by convolutional neural networks (CNNs) are used to extract molecular image features, while decoders based on recurrent neural networks (RNNs) or Transformer generate string-based molecular representations through autoregressive next token prediction. Furthermore, considering the inherent graph properties of molecular structures, previous method \cite{Morin_2023_ICCV} has integrated graph neural networks (GNNs) into the recognition pipeline. The GNN-based approaches generally require preprocessing steps for component detection or segmentation, and typically face challenges in handling stereochemical molecules and Markush structure molecules.

\subsection{Reinforcement Learning and Reasoning}

The integration of Reinforcement Learning (RL) \cite{ouyang2022RLHF, schulman2017PPO} and Chain-of-Thought (CoT) \cite{wei2022chain} have proven effective in enhancing Large Language Models (LLMs) reasoning performance. CoT prompting enables LLMs to decompose complex problems into intermediate reasoning steps, making the problem-solving process more transparent and systematic. While RL can help LLMs to recognize which reasoning processes are correct through reward-based optimization. Popular methods include Direct Preference Optimization (DPO) \cite{rafailov2023DPO}, which has been widely adopted for training reasoning-focused models. More recently, advanced techniques such as GRPO \cite{shao2024deepseekmath} have been introduced, which lead to more stable and effective training dynamics.

\section{Method}

\subsection{OCSR-Specific Image Captioning Pre-training}

Most existing public OCSR datasets contain only chemical structure images and their corresponding SMILES annotations. To leverage these data, a natural approach is to treat the SMILES text as a special description of the chemical structure image, thereby utilizing image captioning techniques to guide model learning. We selected MolParser-7M \cite{fang2024molparser} as our pre-training dataset due to its substantial scale and comprehensive coverage of diverse data sources and image styles. However, the inherent limitations of the original SMILES notation prevent it from describing the numerous Markush structures present in this dataset. To address this limitation, we implemented a simple extension to the SMILES notation system, named SMILES-M. Fig. 2 illustrates the different types of structural variations in Markush structures and their corresponding extended representation methods.

\subsection{Towards More Powerful OCSR Model through Multi-Granularity Learning}

MolScribe \cite{MolScribe} introduces two additional tasks alongside autoregressive SMILES text generation: chemical bond classification and atom localization. In MolScribe, these three tasks are jointly optimized in a collaborative way. Its experimental results demonstrate that by using the predicted types of chemical bonds and the predicted coordinates of atoms to refine the final molecular structures, the accuracy of OCSR can be improved. We further investigated the impact of these two auxiliary tasks on the OCSR task.

\subsubsection{Atoms as Queries}
After processing SMILES text through the tokenizer, the resulting tokens can be categorized into atom tokens and non-atom tokens. Atom tokens correspond to atoms or functional groups within the molecule, while non-atom tokens represent auxiliary information such as brackets indicating the start and end positions of groups, numbers denoting ring connections, and symbols representing chemical bonds or charges. The queries corresponding to atom tokens contain all the information required for the model to predict the respective atoms. We use these queries as inputs for both the chemical bond head and the coordinate head, as shown in Fig. 3.

\subsubsection{Chemical Bond Classification}
To predict the chemical bond $\bm{b}$ from the $i$-th atom $\bm{a_i}$ to the $j$-th atom $\bm{a_j}$, following MolScribe, we simply concatenate the hidden states of these two atom queries and feed them into a chemical bond head for classification, as:
$$
\hat{p}_{i \rightarrow j}\left(\bm{b}\right)={\boldsymbol{\phi}_{bond}\left(\bm{h}_i \oplus \bm{h}_j\right)},
$$
where $\bm{h_i}$ and $\bm{h_j}$ are the hidden states of $\bm{a_i}$ and $\bm{a_j}$, respectively. $\oplus$ indicates the operation of vector vertical concatenation. The chemical bond head $\phi_{bond}$ is implemented as a 2-layer MLP network followed by a softmax.

In practice, we find that SMILES text generation and chemical bond classification exhibit strong synergy, and jointly optimizing these two tasks can effectively improve the training performance of both. More detailed quantitative comparisons can be found in the Main Results and Ablation Study section.

\begin{figure}[t]
    \centering
    \includegraphics[width=1\linewidth]{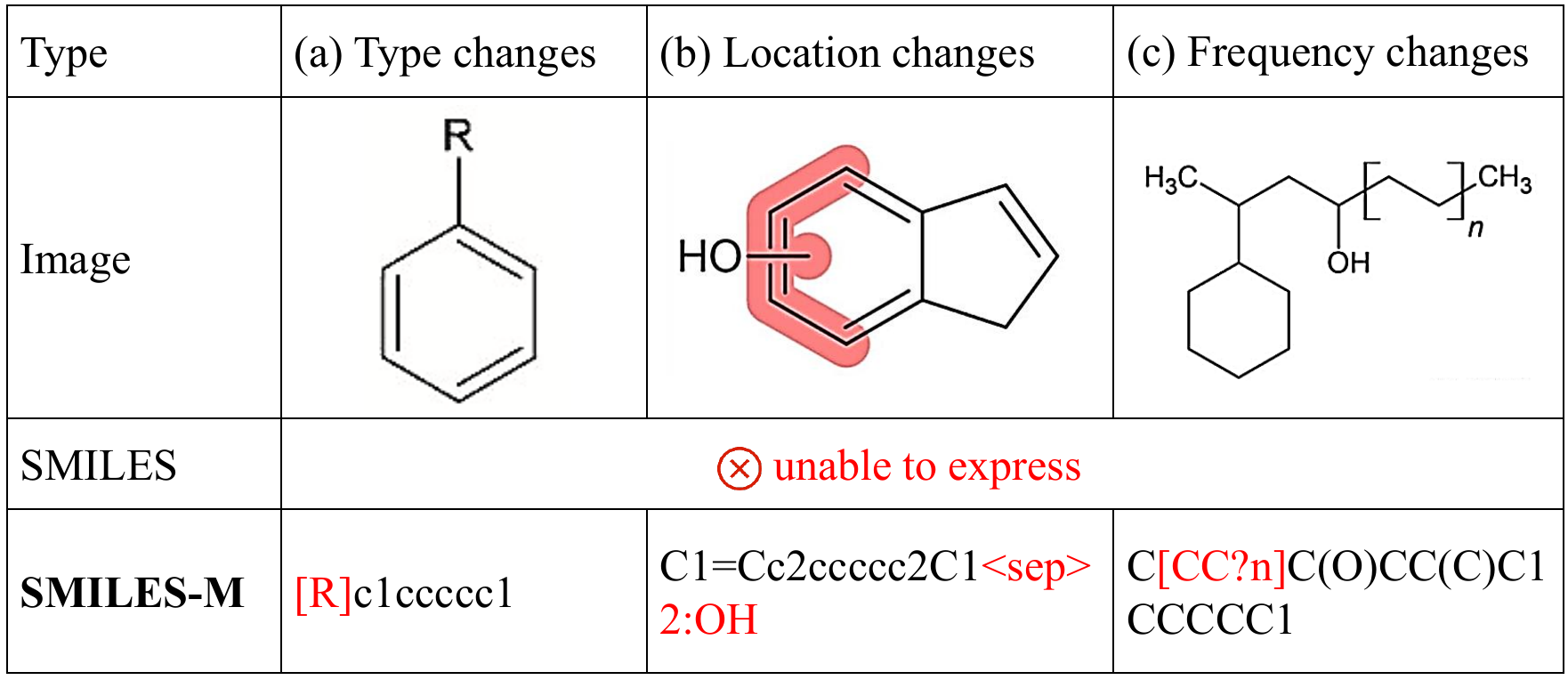}
    \caption{\textbf{Examples of how SMILES-M express Markush structures.} Our SMILES-M can deal with all types of Markush structures, including type changes, location changes, and frequency changes.}
    \label{fig:SMILES-M}
\end{figure}

\begin{figure*}[t]
\centering
\includegraphics[width=0.8\textwidth]{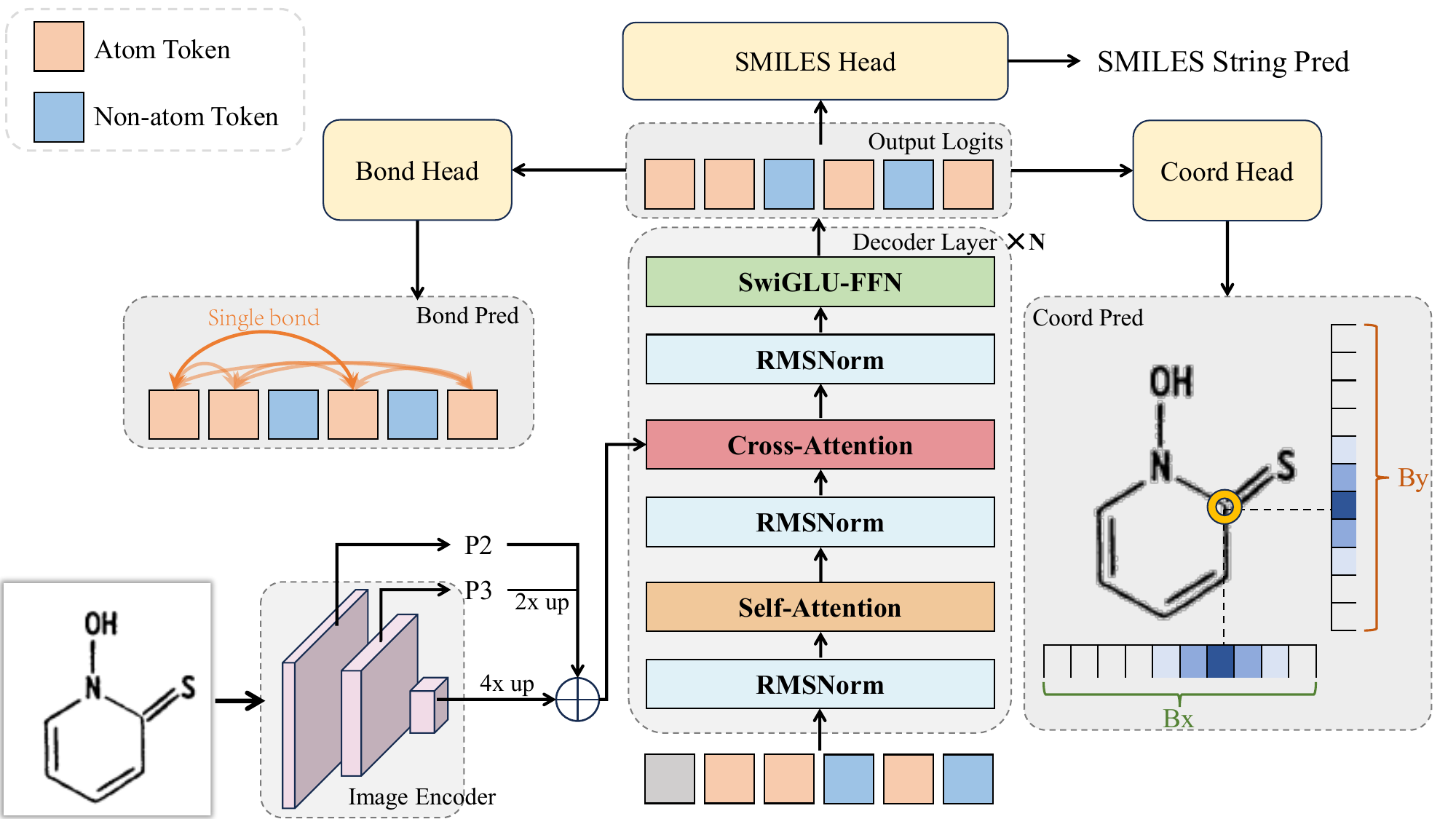}
\caption{\textbf{Overall pipeline of Molsight.} Given a chemical structure image, the image encoder extracts and fuses multi-level image features, which will be fed into the SMILES decoder with previous SMILES tokens to predict the next SMILES token. The SMILES head maps the output logits into SMILES vocabulary space, while the bond head and the coord head predict the chemical bond type and location of each atom token, respectively. The residual connections are omitted in this figure.}
\label{fig:pipeline}
\end{figure*}

\subsubsection{Atom Localization}
Following SimCC \cite{li2022simcc}, we treat the coordinate prediction problem as a one-dimensional classification task for keypoints in both horizontal and vertical directions, which alleviates the localization accuracy issues caused by low-resolution feature maps while maintaining processing speed. MolScribe explicitly inserts coordinate tokens into SMILES sequences for autoregressive learning, but this approach has two significant limitations: (1) As a classification problem, the correct coordinate label is one-hot encoded, and cross-entropy loss is utilized, meaning all incorrect coordinates receive equal penalties except for the single correct coordinate point. However, it would be more reasonable for model predictions closer to the correct coordinates to receive lower penalties. (2) This method increases the target SMILES sequence length by nearly threefold, making it more challenging for the model to handle long sequences. Our MolSight continues to extract coordinate information based on atomic queries, achieving higher localization accuracy without increasing sequence length.

To address the first problem mentioned above, it is necessary to introduce a loss function suitable for coordinate classification tasks. Following RTMO \cite{lu2024rtmo}, we employ maximum likelihood estimation (MLE) to optimize the parameters $\theta$ of the coordinate head. The true coordinate $\mu$ is assumed to follow a Laplace distribution centered at the annotated values $\mu_g$. We use the negative log-likelihood function as our loss function, for the $k$-th atom:
$$
\begin{aligned}
\mathcal{L}_{coord} &= - log \left[ P\left(\mu=\mu_g \mid \theta \right) \right] \\
&= - log \left[ \sum_{i=1}^{B} P\left(\mu_g \mid x_i, \theta\right) \hat{p_k}\left(x_i \mid \theta \right) \right],
\end{aligned}
$$
where
$$
\begin{aligned}
P\left(\mu_g \mid x_i, \theta\right) &= \frac{1}{2\hat{b}} e^{-\frac{\left|\mu_g - x_i\right|}{\hat{b}}}, \\
\hat{p_k}\left(x_i \mid \theta \right) &= 
\frac{e^{\boldsymbol{h}_k \cdot \boldsymbol{\phi}_{coord}\left(\boldsymbol{P} \boldsymbol{E}\left(x_i\right)\right)}}{\sum_{j=1}^{B} e^{\boldsymbol{h}_k \cdot \boldsymbol{\phi}_{coord}\left(\boldsymbol{P} \boldsymbol{E}\left(x_j\right)\right)}},
\end{aligned}
$$
the scale parameter $\hat{b}$ of the Laplace distribution is additionally predicted by the coordinate head, it reflects the model's uncertainty regarding the current coordinate prediction results, and can serve as a reference for filtering subsequent molecular structure prediction results. The positional encoding function $\boldsymbol{P} \boldsymbol{E}$ is the same as the one from Transformer \cite{vaswani2017attention}, $B$ is a hyperparameter denotes the number of coordinate bins, and $x_i$ and $x_j$ are discrete horizontal/vertical coordinate values picked from $linspace(0, 1, B)$.

In our early experiments, we found that incorporating the atom localization task into our training pipeline negatively impacted the optimization of the other two tasks. Additionally, directly using the hidden states from the SMILES decoder for coordinate prediction failed to achieve satisfactory accuracy. To address these issues, we introduced two additional Transformer layers to process the atom queries before feeding them to the coordinate head. Furthermore, the coordinate head is optimized independently from the other two tasks.

\begin{figure*}[t]
\centering
\includegraphics[width=0.7\textwidth]{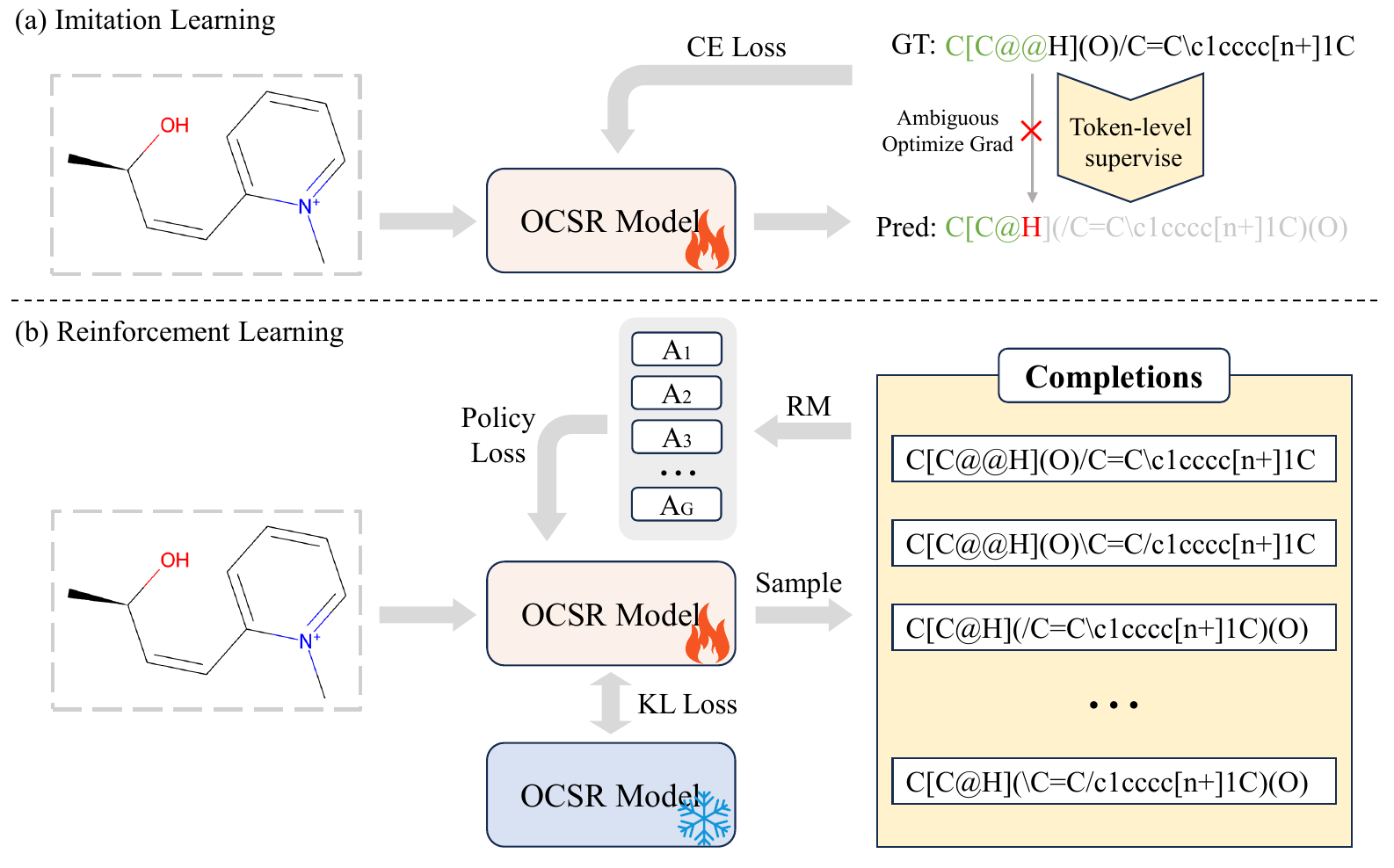}
\caption{\textbf{Comparison of Training Paradigms.} (a) \textbf{Imitation Learning:} due to the diversity of correct SMILES text, token-level optimization may cause ambiguous optimize direction. (b) \textbf{Reinforcement Learning:} multiple completions are sampled at once, then scored on Tanimoto similarity and structural consistency. We use GRPO to achieve this trajectory-level optimization.}
\label{fig:rl}
\end{figure*}

\subsection{Semantic-aware Reinforcement Learning}
SMILES text follows explicit syntactic rules (such as parentheses and ring number pairing), and generation errors are costly—a single incorrect character can cause the entire SMILES invalid, leading to failure in downstream tasks (e.g., preventing use in chemical modeling). Therefore, the metric of primary concern is typically not token-level accuracy, but rather the overall accuracy of SMILES strings (i.e., whether they are completely correct), which is not differentiable and cannot be directly used for gradient optimization. Furthermore, a molecule may have multiple valid SMILES texts, particularly for stereoisomers, where aromatic rings, cis-trans configurations, and chiral centers can be expressed in various correct formats. In such cases, relying solely on imitation learning has inherent limitations, as inconsistent annotation information leads to conflicting optimization objectives.

As shown in Fig. 4, a more direct approach is to allow the model to explore more possible SMILES texts and then apply RL to directly optimize the overall correctness/similarity
\begin{minipage}[c][.525\textheight][c]{\linewidth}
\begin{algorithm}[H]
\caption{Reward Calculation for Molecular Structure}
\label{alg:combined_reward}
\textbf{Input}: Pred SMILES $S_{pred}$, Ground Truth SMILES $S_{gt}$ \\
\textbf{Parameter}: Tanimoto weight $w_t = 0.4$, Stereochemistry weight $w_s = 0.6$ \\
\textbf{Output}: Total Reward $R_{total}$

\begin{algorithmic}[1]
\STATE $S'_{pred} \leftarrow \text{ExpandAbbreviations}(S_{pred})$
\STATE $S'_{gt} \leftarrow \text{ExpandAbbreviations}(S_{gt})$
\STATE Initialize $R_{total} \leftarrow \emptyset$

\FOR{each pair $(s_{p}, s_{g}) \in (S'_{pred}, S'_{gt})$}
    \STATE $r_{tanimoto} \leftarrow 0.0$
    \STATE $r_{stereo} \leftarrow 0.0$
    
    \STATE $M_p \leftarrow \text{MoleculeFromSMILES}(s_p)$
    \STATE $M_g \leftarrow \text{MoleculeFromSMILES}(s_g)$

    \IF{$M_p$ is valid AND $M_g$ is valid}
        
        \STATE \COMMENT{-- Tanimoto Similarity Reward --}
        \STATE $FP_p \leftarrow \text{Fingerprint}(M_p)$
        \STATE $FP_g \leftarrow \text{Fingerprint}(M_g)$
        \STATE $r_{tanimoto} \leftarrow \text{TanimotoSimilarity}(FP_p, FP_g)$
        
        \STATE \COMMENT{-- Stereochemistry Reward --}
        \IF{$\text{InChIKey}(M_p) = \text{InChIKey}(M_g)$}
            \STATE $r_{stereo} \leftarrow 1.0$ \COMMENT{Exact match}
        \ELSIF{$\text{AtomCount}(M_p) = \text{AtomCount}(M_g)$}
            \STATE $r_{stereo} \leftarrow 0.3$ \COMMENT{Similar structure}
        \ELSE
            \STATE $r_{stereo} \leftarrow 0.1$
        \ENDIF
    \ENDIF
    
    \STATE $r_{combined} \leftarrow w_t \times r_{tanimoto} + w_s \times r_{stereo}$
    \STATE Append $r_{combined}$ to $R_{total}$
\ENDFOR

\STATE \textbf{return} $R_{total}$
\end{algorithmic}
\end{algorithm}
\end{minipage}
metrics. For stereoisomers, this direct optimization of chemical semantic correctness can also guide the model to attend to stereochemical markers, bond orientations, and other critical details in molecular images, thereby improving the model's ability to recognize stereoisomeric molecules.

We choose GRPO as the RL algorithm for MolSight, the loss function is defined as follows:
\begin{footnotesize}
$$ 
\begin{aligned}
&\mathcal{L}_{\mathrm{GRPO}}(\theta)= \\
&-\frac{1}{G} \sum_{i=1}^G \frac{1}{\left|o_i\right|} \sum_{t=1}^{\left|o_i\right|}\left[\frac{\pi_\theta\left(o_{i, t} \mid o_{i, <t}\right)}{\left[\pi_\theta\left(o_{i, t} \mid o_{i, <t}\right)\right]_{\mathrm{no\ grad}}} \hat{A}_{i, t} -\beta \mathbb{D}_{\mathrm{KL}}\left[\pi_\theta \| \pi_{\mathrm{ref}}\right]\right],
\end{aligned}
$$
\end{footnotesize}
where ${o_1, o_2, ..., o_G}$ are a group of sampled outputs, the advantage $\hat{A}_{i, t}$ is computed using the normalized reward within the group.

The reward of SMILES consists of two parts: tanimoto similarity reward and stereochem reward. The processing of chemical information is done by RDKit \cite{landrum2016rdkit}. The details of our reward function are shown in Alg. 1.

\begin{table*}[t]
\centering
\renewcommand{\arraystretch}{1.5} 
\setlength{\tabcolsep}{1.3mm}{

\begin{tabular}{l|cccc|ccc|cccc|ccc}
\hline
\multirow{2}{*}{Method} & \multicolumn{4}{c|}{USPTO} & \multicolumn{3}{c|}{UoB} & \multicolumn{4}{c|}{CLEF} & \multicolumn{3}{c}{JPO} \\ \cline{2-15} 
 & graph & stereo & exact & tani & graph & exact & tani & graph & stereo & exact & tani & graph & exact & tani \\ \hline
OSRA-2.2.1 & 86.5 & 79.0 & 83.5 & 92.9 & 76.1 & 75.5 & 88.4 & 86.2 & \textbf{89.3} & 83.3 & 88.4 & 51.8 & 50.7 & 69.1 \\
MolVec-0.9.8 & 91.4 & \underline{82.8} & 88.1 & 95.8 & 80.7 & 80.2 & 91.5 & 84.1 & \underline{89.1} & 82.8 & 88.5 & \textbf{68.9} & \underline{66.4} & 85.0 \\
Imago-2.0 & 89.2 & 77.4 & 87.2 & 94.5 & 58.1 & 57.6 & 77.1 & 65.1 & 45.1 & 60.3 & 84.1 & 40.7 & 40.4 & 59.6 \\
DECIMER-2.7.1 & 61.5 & 43.8 & 58.4 & 92.4 & 87.1 & 86.4 & \textbf{96.2} & 80.4 & 76.0 & 73.8 & 89.1 & 40.4 & 39.3 & 86.4 \\
MolGrapher$\dagger$ & 82.9 & 10.3 & 65.7 & 93.3 & 84.9 & 84.2 & 93.6 & 71.1 & 10.4 & 52.5 & 86.7 & 61.1 & 52.2 & 76.3 \\
MolScribe & \textbf{94.6} & 69.0 & \underline{88.4} & \textbf{97.5} & \textbf{88.2} & \textbf{87.4} & \underline{96.0} & \underline{91.8} & 76.8 & \textbf{85.5} & \textbf{90.5} & 60.0 & 57.6 & \underline{88.3} \\
\textbf{MolSight (ours)} & \underline{94.0} & \textbf{85.1} & \textbf{92.0} & \underline{97.4} & \underline{87.9} & \underline{87.1} & 95.9 & \textbf{92.2} & 76.0 & \underline{84.9} & \underline{90.1} & \underline{68.7} & \textbf{66.7} & \textbf{90.7} \\ \hline
\end{tabular}
}
\caption{\textbf{Performance comparison with existing SOTA methods on real data benchmarks.} $\dagger$: the stereo-capable variant of MolGrapher was used, \textbf{graph}: recognition accuracy ignoring stereoisomers, \textbf{stereo}: recognition accuracy on stereochemical molecules, \textbf{exact}: exact recognition accuracy, \textbf{tani}: Tanimoto Coefficient.}
\label{table1}
\end{table*}

\begin{table}[]
\centering
\renewcommand{\arraystretch}{1.3}
\begin{tabular}{m{4cm}cc}
\hline
\textbf{Setting} & \textbf{graph} & \textbf{stereo} \\ \hline
\multicolumn{3}{l}{\textit{Get stereo from SMILES}} \\ \hline
SMILES & 93.7 & 80.6 \\
+ edge & 94.2 & 80.9 \\ \hline
\multicolumn{3}{l}{\textit{Get stereo from edges \& coords}} \\ \hline
Joint Training & 43.8 & 10.6 \\
Separate Training & 94.2 & 29.4 \\
+ extra decoder layers & & \\
\quad w/ L1 Loss & 94.2 & 82.4 \\
\quad w/ MLE Loss & \textbf{94.2} & \textbf{83.7} \\ \hline
\end{tabular}
\caption{\textbf{Ablation on proposed training tasks.} The results were obtained from USPTO dataset. 
}
\label{table2}
\end{table}

\section{Experiments}
\subsection{Implementation Details}
MolSight employs an encoder-decoder architecture. We select EfficientViT (L1 version with approximately 53M parameters) \cite{cai2023efficientvit} as the image encoder due to its lightweight convolutional architecture and linear attention mechanism. The decoder adopts a 6-layer standard Transformer architecture with several common improvements including RoPE \cite{su2024roformer}, SwiGLU \cite{shazeer2020glu}, and RMSNorm \cite{zhang2019root}. The decoder parameters are randomly initialized rather than using pre-trained language models for initialization, as there exists a significant vocabulary mismatch between standard language models and SMILES text.

First, we pre-train the model for 2 epochs on a dataset containing only SMILES annotations using next token prediction. Subsequently, we jointly train the model for 10 epochs on both SMILES text generation and chemical bond classification (experimental results demonstrate that extending training to 30 epochs can further improve model performance), while adapting the model with a coordinate head. Finally, we further optimize the model for 2 epochs using RL methods.

\subsection{Dataset and Evaluation metric}
\subsubsection{Dataset}
During the pre-training stage, we train our model on the MolParser-7M dataset from \cite{fang2024molparser} for SMILES text generation tasks. In the fine-tuning phase, we employ the PubChem-1M and USPTO-680K from \cite{MolScribe} for multi-granularity learning. During the post-training phase, we enhance the model using reinforcement learning on our self-collected stereochemical molecular dataset, named Stereo-200K. For evaluation, we report model performance on four classical real-world datasets (USPTO \cite{filippov2009optical}, Maybridge UoB \cite{sadawi2012chemicalUoB}, CLEF-2012 \cite{piroi2010clef}, and JPO \cite{fujiyoshi2011robustJPO}) and four synthetic datasets (Staker \cite{staker2019molecular}, ChemDraw, Indigo, and Stereo-2K). Both the Stereo-200K and Stereo-2K datasets are introduced for the first time in this work.

\subsubsection{Metric}
For molecular recognition performance, we use exact match accuracy and average molecular fingerprint similarity (i.e., Tanimoto Coefficient) as evaluation metrics. For atom localization, we employ the Object Keypoint Similarity (OKS) metric to assess the performance, as:
$$
OKS = \frac{1}{N_{kpts}} \sum_{n=1}^{N_{kpts}} e^{- \frac{d^2_n}{2s^2}},
$$
where $N_{kpts}$ is the number of keypoints, $d_n$ is the Euclidean distance between the n-th predicted location and the ground-truth location, and $s$ is a scale factor. 

\subsection{Main Results and Ablation Study}
\subsubsection{Experiments on Real Data Benchmarks}
Our first objective is to develop an plain OCSR model that achieves advanced performance on in-domain data by generating SMILES text alone, thereby facilitating efficient fine-tuning for users on specific target datasets.

\begin{table}[ht!]
\centering
\renewcommand{\arraystretch}{1.5}
\setlength{\tabcolsep}{1.2mm}{ 
\begin{tabular}{ll|ccc|ccc}
\hline
\multicolumn{2}{l|}{\multirow{2}{*}{\begin{tabular}[c]{@{}l@{}}Training\\ Strategy\end{tabular}}} &
  \multicolumn{3}{c|}{Stereo-2k} &
  \multicolumn{3}{c}{CLEF} \\ \cline{3-8} 
\multicolumn{2}{l|}{}          & graph & stereo & tani & graph         & stereo & tani \\ \hline
\multicolumn{2}{l|}{SFT}       & 95.4  & 80.1   & 98.5 & 92.2          & 71.0   & 90.0 \\ \hline
\multicolumn{1}{l|}{\multirow{3}{*}{\begin{tabular}[c]{@{}l@{}}SFT\\ +RL\end{tabular}}} &
  Tanimoto &
  \textbf{97.2} &
  81.8 &
  99.2 &
  92.2 &
  74.3 &
  \textbf{90.1} \\
\multicolumn{1}{l|}{} & Stereo & 96.6  & 86.5   & 98.9 & \textbf{92.6} & 78.4   & 90.0 \\
\multicolumn{1}{l|}{} &
  Weighted &
  96.9 &
  \textbf{87.1} &
  \textbf{99.3} &
  91.8 &
  \textbf{80.6} &
  90.0 \\ \hline
\end{tabular}
}
\caption{\textbf{Ablation on different training strategies.} Further fine-tuning the model with RL led to additional performance gains. \textbf{Tanimoto}: only use Tanimoto Similarity Reward, \textbf{Stereo}: only use Stereochemistry Reward, \textbf{Weighted}: weighted reward function as shown in Alg. 1.}
\label{table3}
\end{table}

Tab. 1 presents a quantitative comparison of our method against current SOTA approaches on four classic real-world benchmarks. The stereochemical recognition accuracy for UoB and JPO datasets is not available since these datasets contain virtually no stereochemical molecules. For fair comparison, both our method and MolScribe were fine-tuned for 30 epochs on the same datasets without using chemical bond information or atomic coordinate information for auxiliary correction.

The results demonstrate that compared to MolScribe, our MolSight model exhibits superior performance on in-domain data (i.e., the USPTO dataset), particularly achieving a stereochemical recognition accuracy of 85.1\% (\textbf{+16.1}\%). Additionally, on the more challenging JPO dataset, MolSight attains an exact recognition accuracy of 66.7\% (\textbf{+9.1}\%) and a graph recognition accuracy of 68.7\% (\textbf{+8.7}\%). Furthermore, Our MolSight is the only method that achieves Tanimoto Coefficients exceeding 90\% across all four benchmarks.

\begin{table*}[t]
\centering
\renewcommand{\arraystretch}{1.2} 
\setlength{\tabcolsep}{1.3mm}{

\begin{tabular}{lcccccccc}
\hline
Methods                         & BBBP     & Tox21    & ToxCast  & SIDER    & ClinTox  & MUV      & HIV      & Bace     \\ \hline
\multicolumn{9}{l}{\textit{Multimodal}}                                                                                 \\
3D InfoMax \cite{stark20223dinfomax}                     & 69.1±1.0 & 74.5±0.7 & 64.4±0.8 & 60.6±0.7 & 79.9±3.4 & 74.4±2.4 & 76.1±1.3 & 79.7±1.5 \\
GraphMVP \cite{liu2021pregraphmvp}                       & 68.5±0.2 & 74.5±0.4 & 62.7±0.1 & 62.3±1.6 & 79.0±2.5 & 75.0±1.4 & 74.8±1.4 & 76.8±1.1 \\
MoleBlend \cite{yu2024multimodalmoleblend}                      & \textbf{73.0±0.8} & \textbf{77.8±0.8} & \textbf{66.1±0.0} & \textbf{64.9±0.2} & \textbf{87.6±0.7} & \textbf{77.2±2.3} & \textbf{79.0±0.8} & \textbf{83.7±1.4} \\ \hline
\multicolumn{9}{l}{\textit{2D graph}}                                                                                   \\
MolCLR \cite{wang2022molclr}                         & 66.6±1.8 & 73.0±0.1 & 62.9±0.3 & 57.5±1.7 & \textbf{86.1±0.9} & 72.5±2.3 & 76.2±1.5 & 71.5±3.1 \\
GraphMAE \cite{hou2022graphmae}                       & \textbf{72.0±0.6} & 75.5±0.6 & 64.1±0.3 & 60.3±1.1 & 82.3±1.2 & 76.3±2.4 & 77.2±1.0 & \textbf{83.1±0.9} \\
Mole-BERT \cite{xia2023molebert}                      & 71.9±1.6 & \textbf{76.8±0.5} & \textbf{64.3±0.2} & \textbf{62.8±1.1} & 78.9±3.0 & \textbf{78.6±1.8} & \textbf{78.2±0.8} & 80.8±1.4 \\ \hline
\multicolumn{9}{l}{\textit{Image}}                                                                                      \\
EfficientViT (ImageNet pretrain) & 63.6±0.8 & 71.3±0.5 & 63.2±0.4 & 60.4±0.4 & 98.0±0.3 & 66.9±1.2 & 73.8±1.2 & 73.9±1.5 \\
EfficientViT (MolSight pretrain) & \textbf{68.0±0.6} & \textbf{75.0±0.4} & \textbf{65.0±0.2} & \textbf{62.6±0.7} & \textbf{98.4±0.9} & \textbf{73.6±1.0} & \textbf{74.1±1.1} & \textbf{76.9±1.6} \\ \hline
\end{tabular}
}
\caption{\textbf{Results on molecular property prediction tasks.} ROC-AUC scores are reported (higher is better). The MolSight pre-training method demonstrates superior performance over the ImageNet \cite{deng2009imagenet} pre-training approach across all evaluated datasets, and comparable to those advanced molecular pre-training methods that are based on 2D graph or multimodal.}
\label{table4}
\end{table*}

\subsubsection{Ablation Study on Training Tasks}
In the second training stage, we introduce two auxiliary learning objectives: chemical bond classification and atom localization. This serves a dual purpose. First, we aim to validate whether adding more supervision information can improve the results of SMILES text generation. Second, these auxiliary tasks provide a novel basis for determining the molecule's stereochemistry. By leveraging the RDKit toolkit, we can infer the complete 3D molecular conformation, including specific chirality and cis-trans isomerism, from the predicted 2D coordinates and chemical bond types.

Tab. 2 presents the ablation study of our proposed training tasks. The results indicate that augmenting the SMILES decoder with a chemical bond head not only enables the model to predict chemical bonds but also slightly improves the accuracy of SMILES text generation.

However, the introduction of the atom localization task complicates the overall optimization. Jointly optimizing all three objectives causes the model to over-prioritize the atom position detection branch, which in turn hinders the convergence of the primary SMILES generation task. This is reflected in the training metrics as a high OKS score but a low token-level SMILES accuracy.

To address this issue, we decoupled the coordinate prediction branch and optimized it separately. This approach effectively prevents the coordinate prediction from interfering with the optimization of other tasks. Nevertheless, directly using the output queries from the SMILES decoder for coordinate prediction did not yield satisfactory localization performance, 
so we added two additional decoder layers into the coordinate prediction branch to further extract positional information from the input image. This modification led to a significant improvement, where the accuracy of the molecular stereochemistry inferred from the predicted chemical bonds and coordinates surpassed that derived directly from the generated SMILES text.

\subsubsection{Impact of Reinforcement Learning}
We further investigated the use of RL post-training to encourage the model to autonomously explore correct SMILES notations, thereby enhancing its accuracy in recognizing stereoisomers. To this end, we constructed the Stereo dataset, which is composed entirely of stereoisomer images. These molecular data were sourced from the PubChem database and include a balanced distribution of chiral and cis-trans isomers, as well as a significant number of structurally similar molecules. We utilized the RDKit toolkit to generate the molecular images with random stylistic variations. Tab. 3 presents a comparison of the model's performance on both in-domain (Stereo-2k) and out-of-domain (CLEF-2012) data, before and after RL-based training on the Stereo-200k dataset.

\subsubsection{Analysis of Transfer Learning Performance}
To validate the transferability of features learned by MolSight, we evaluated our model on downstream tasks from the MoleculeNet \cite{wu2018moleculenet}, one of the most widely used benchmarks for molecular property prediction. Our evaluation procedure was plain: First, each molecule was rendered into a 2D image using the Indigo toolkit \cite{pavlov2011indigo}. Next, the MolSight image encoder, with its parameters frozen, was employed to extract features from each image. These image features, after undergoing global average pooling, were then fed into a trainable MLP probe for the final classification. Following standard evaluation protocols for this benchmark, all datasets were partitioned using the scaffold split method, which groups molecules by their structural backbone to provide a more rigorous test of model generalization. Tab. 4 reports the ROC-AUC scores on all eight classification tasks. We report the mean and standard deviation across 5 random seeds (from 0 to 4).

\section{Conclusion}
In this paper, we introduced MolSight, a comprehensive framework designed to address the challenges of Optical Chemical Structure Recognition (OCSR), particularly in accurately interpreting stereochemistry. Our novel three-stage training paradigm, which synergizes large-scale pre-training, multi-granularity fine-tuning, and reinforcement learning, proves highly effective. Extensive experiments demonstrate that MolSight achieves state-of-the-art performance across several challenging benchmarks. Furthermore, the strong transfer learning performance on downstream molecular property prediction tasks validates the robustness of the features learned by our model, highlighting its potential for broader applications in automated chemical data analysis.

\clearpage
\appendix

\section{Supplementary Material}

\subsection{Additional Results}

\subsubsection{Impact of different drawing styles}
To evaluate the impact of different drawing styles on the recognition accuracy of MolSight, we redrew molecules from the USPTO patent dataset using the ChemDraw and Indigo toolkits. Table 5 presents the model's performance on these datasets. Overall, the model performs slightly better on the synthetic data compared to the original patent images. Furthermore, it achieves slightly higher accuracy on images rendered by Indigo than on those by ChemDraw. A possible explanation for these findings is that the synthetic data has a lower noise level, and critically, the model itself was trained using data produced by Indigo.
\begin{table}[H]
\centering
\renewcommand{\arraystretch}{1.3}
\begin{tabular}{l|cccc}
\hline
Data Source & graph & stereo & exact & tani \\ \hline
Patent      & 94.0  & 85.1   & 92.0  & 97.4 \\
ChemDraw    & 95.3  & 84.0   & 92.5  & 98.7 \\
Indigo      & 96.9  & 86.9   & 94.1  & 99.0 \\ \hline
\end{tabular}
\caption{\textbf{Performance comparison on different drawing styles.} The images in these three datasets depict the same batch of molecules, respectively.
}
\label{table5}
\end{table}

\subsubsection{Impact of image perturbation \& degradation}
To assess MolSight's robustness, we tested its performance on the perturbed datasets(i.e., USPTO$_p$, UoB$_p$, CLEF$_p$, and JPO$_p$) and the low-resolution Staker dataset \cite{staker2019molecular}, as shown in Tab. 6. The images in the perturbed datasets were generated by applying random rotations within [-5°, 5°] and random xy-shearing with a factor between [-0.1, 0.1]. Each original image was perturbed five times. This procedure was designed to detect potential overfitting to the public datasets. The Staker dataset contains 50,000 low-resolution images synthesized with the Indigo toolkit, which simulates low-quality images found in real-world data through an aggressive downsampling strategy. The results demonstrate that MolSight exhibits greater robustness to both image perturbation and degradation compared to rule-based methods.
\begin{table}[t]
\centering
\renewcommand{\arraystretch}{1.3}
\setlength{\tabcolsep}{1mm}{
\begin{tabular}{l|ccccc}
\hline
Method & USPTO$_p$ & UoB$_p$ & CLEF$_p$ & JPO$_p$ & Staker \\ \hline
OSRA-2.2.1 &
  \begin{tabular}[c]{@{}c@{}}6.7\\ {\small \textcolor{blue}{(-76.8)}}\end{tabular} &
  \begin{tabular}[c]{@{}c@{}}66.9\\ {\small \textcolor{blue}{(-8.6)}}\end{tabular} &
  \begin{tabular}[c]{@{}c@{}}18.6\\ {\small \textcolor{blue}{(-64.7)}}\end{tabular} &
  \begin{tabular}[c]{@{}c@{}}31.3\\ {\small \textcolor{blue}{(-19.4)}}\end{tabular} &
  0.0 \\ \hline
MolVec-0.9.8 &
  \begin{tabular}[c]{@{}c@{}}29.6\\ {\small \textcolor{blue}{(-58.5)}}\end{tabular} &
  \begin{tabular}[c]{@{}c@{}}74.1\\ {\small \textcolor{blue}{(-6.1)}}\end{tabular} &
  \begin{tabular}[c]{@{}c@{}}43.7\\ {\small \textcolor{blue}{(-39.1)}}\end{tabular} &
  \begin{tabular}[c]{@{}c@{}}52.1\\ {\small \textcolor{blue}{(-14.3)}}\end{tabular} &
  0.7 \\ \hline
Imago-2.0 &
  \begin{tabular}[c]{@{}c@{}}42.9\\ {\small \textcolor{blue}{(-44.3)}}\end{tabular} &
  \begin{tabular}[c]{@{}c@{}}39.6\\ {\small \textcolor{blue}{(-18.0)}}\end{tabular} &
  \begin{tabular}[c]{@{}c@{}}40.6\\ {\small \textcolor{blue}{(-19.7)}}\end{tabular} &
  \begin{tabular}[c]{@{}c@{}}23.1\\ {\small \textcolor{blue}{(-17.3)}}\end{tabular} &
  0.0 \\ \hline
MolSight &
  \begin{tabular}[c]{@{}c@{}}92.3\\ {\small \textcolor{red}{(+0.3)}}\end{tabular} &
  \begin{tabular}[c]{@{}c@{}}86.0\\ {\small \textcolor{blue}{(-1.1)}}\end{tabular} &
  \begin{tabular}[c]{@{}c@{}}85.5\\ {\small \textcolor{red}{(+0.6)}}\end{tabular} &
  \begin{tabular}[c]{@{}c@{}}68.7\\ {\small \textcolor{red}{(+2.0)}}\end{tabular} &
  82.1 \\ \hline
\end{tabular}
}
\caption{\textbf{Performance comparison on perturbed benchmarks.} We reported the exact recognition accuracy for all datasets. Numbers in parentheses indicate the performance change from the original datasets (red for improvement, blue for decline).}
\label{table6}
\end{table}

\subsubsection{Reward changing with training steps.}
In Figure 5, we show the reward curve during RL training process. Under the default training setting (i.e. weighted reward function), the average reward per batch increased from 0.84 to 0.92. We also provided the reward curve using Tanimoto Similarity Reward or Stereochemistry Reward alone.

\begin{figure*}[t]
\centering
    \includegraphics[width=0.8\textwidth]{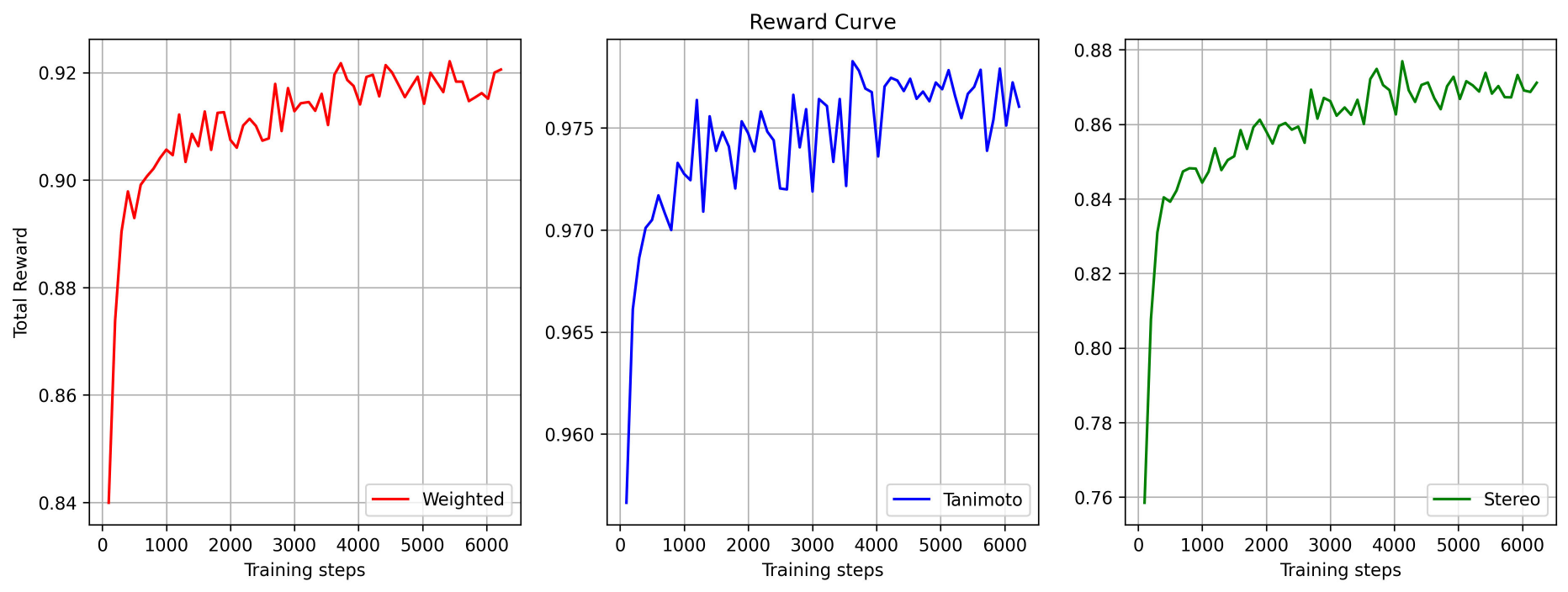}
    \caption{\textbf{Reward curve during the training process.} The total reward rises steadily throughout post training. \textbf{Left}: training with the weighted reward function as shown in Alg. 1, \textbf{Middle}: training with the Tanimoto Similarity Reward alone, \textbf{Right}: training with the Stereochemistry Reward alone.}
    \label{fig:reward_curve}
\end{figure*}

\subsection{More Implementation Details}

\subsubsection{Model details.}
The input images are preprocessed to a resolution of 512×512. This is achieved by cropping whitespace, resizing the image, and then padding the shorter side to create a square format. During the training stage, we apply a series of data augmentation techniques, including random rotation, downsampling, blurring, and the addition of Gaussian and salt-and-pepper noise. For the synthetic dataset, we further enhance data diversity by employing randomized rendering options and performing molecular subgroup replacement.

We use the AdamW \cite{loshchilov2017decoupled} optimizer ($\beta_1=0.9$, $\beta_2=0.999$) and a linear learning rate warmup. In the first training stage, we train the model with a batch size of 512, and a base learning rate (lr) of $4e^{-4}$, and a weight decay of $1e^{-4}$. we decrease the lr by a factor of 10 for the encoder parameters. The warmup ratio is set to 0.02,  then the lr decays cosinely to 0.We also use the label smoothing with confidence=0.9.
In the second stage, the batch size is set to 256, we first set the base lr of the edge head to $4e^{-4}$, and finetune the whole model on PubChem-1M and USPTO-680K. Then we set the base lr of the coordinate head to $4e^{-5}$, and freeze other parameters of the model, and finetune the coordinate head on PubChem-1M for 2 epochs, the number of coordinate bins is 128.
In the third stage, we use LoRA \cite{hu2022lora} with rank=8 and alpha=16 to optimizing the decoder. The batch size is 64, the base lr is $1e^{-4}$. During the sequence sampling, we sample 4 completions per image, with temperature=1.0 and topk=10.

We implement MolSight on Ubuntu with PyTorch, training and inference are performed on a sigle node equipped with an Intel Xeon Silver 4210R CPU and four NVIDIA GeForce RTX 3090/4090 GPUs.

\subsubsection{SMILES-M details.}
Existing Markush-capable line notations of chemical structures~\cite{fang2024molparser, morin2025markushgrapher} are typically based on SMILES, with additional XML-like extensions to express Markush groups. Inspired by prior works, Our SMILES-M is designed to adapt the majority of Markush structure molecules with minimal modifications to SMILES, while maintaining good readability and avoiding significant increases in sequence length. The general format of SMILES-M is as follows:
{\scriptsize 
$$MAIN\_PART<sep>EXTENSION_{1}...<sep>EXTENSION_{n}$$
}where the EXTENTION suffixes are used to denote position-variable groups, which can appear at any position on one or more rings. The format of an EXTENTION is defined as follows:
{\scriptsize 
$$RING\_INDEX_1,...,RING\_INDEX_m:GROUP\_NAME$$
}where multiple RING\_INDEX indicators correspond to multiple possible rings. 

If there are no position-variable groups in the molecule, then its SMILES-M contains no delimiters or suffixes (i.e, only the SMILES-style MAIN\_PART remains). For R-groups or frequency-variable groups, SMILES-M uses square brackets to wrap their group names, and the question mark is used to separate the group name from its frequency variable.

Although the structure of SMILES-M is concise, during the usage process, we found that this expression method still has some flaws. For example, when multiple rings are placed together, it is not easy to determine the index of each ring at this time

\subsubsection{Stereo-200k details.}
Data Collection Pipeline: The molecular data were retrieved from the first 2 million compounds in the PubChem database (i.e., Compound CID ranging from 1 to 2,000,000). We did not expand the search scope of the PubChem database because we found that molecules with adjacent CIDs often share similar structures, which helps the model learn to distinguish between similar molecular images. Chiral molecules were filtered by verifying the presence of the `@' character in their SMILES strings, while cis-trans isomers were identified through checking for the presence of `$/$' or `$\backslash$' characters in their SMILES strings.

Plotting Workflow: To introduce stylistic diversity, one style was randomly selected from 5 predefined rdkit drawing options: [``classic'', ``slight\_blue\_bg'', ``light\_gray\_bg'', ``with\_atom\_indices'', ``thicker\_lines''], with corresponding sampling weights of [60, 10, 10, 10, 10].

\subsection{Visualizations}
Fig. 6 displays the attention maps from MolSight's Transformer decoder during the SMILES string decoding process. By visualizing these maps as heatmaps, we can observe the image regions that the model attends to for various character types, including atoms, chemical bonds, and auxiliary symbols. This quantitative result demonstrates a strong alignment between the attended regions and the characters' ground-truth locations. Furthermore, as shown in Figure 7, we also visualize the heatmaps of atomic coordinates, which are predicted by the coordinate prediction head for each atom character.

\begin{figure*}[t]
\centering
\includegraphics[width=1\textwidth]{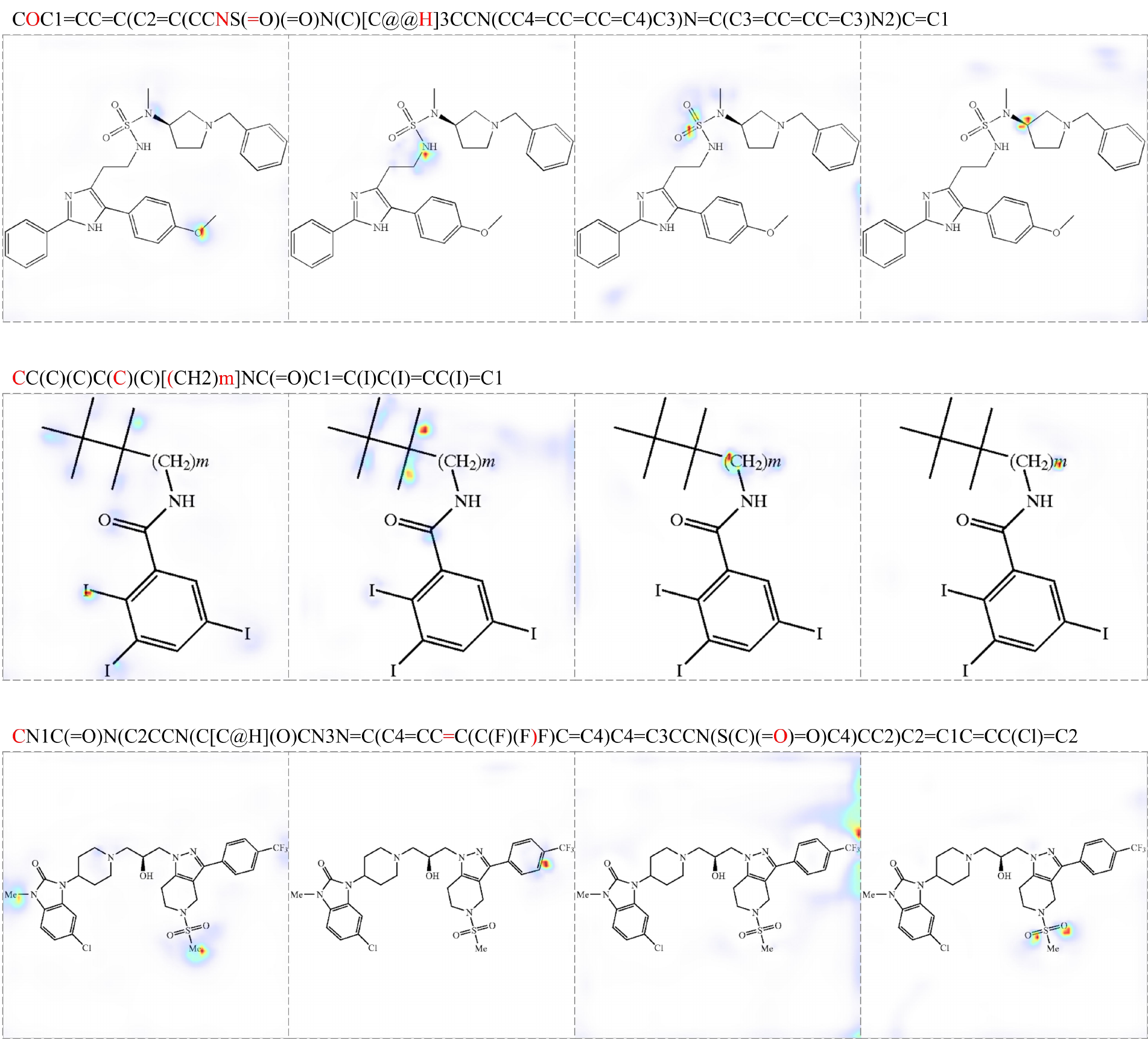}
\caption{\textbf{Visualizations of attention map.} Here we showed the attention map from the 3-th decoder layer. Attention maps from different heads are averaged. We use \textcolor{red}{red} color to indicate the current character.}
\label{fig:attnmap}
\end{figure*}

\begin{figure*}[t]
\centering
\includegraphics[width=\textwidth]{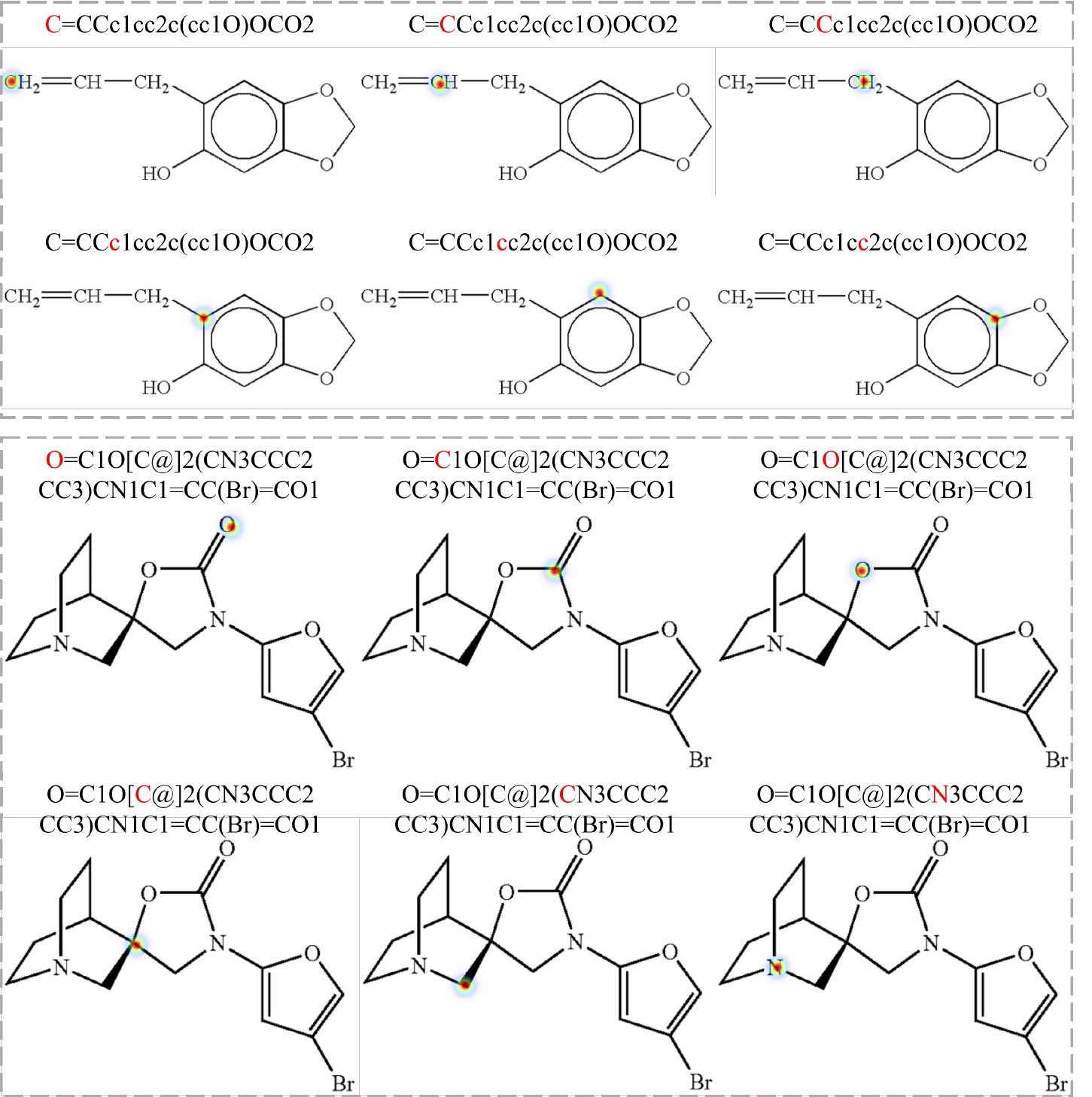}
\caption{\textbf{Visualizations of coordinate heatmap.} We use \textcolor{red}{red} color to indicate the current character. The heatmaps are from the first 6 atoms in the SMILES string.}
\label{fig:coordmap}
\end{figure*}

\clearpage
\ifdefined\aaaianonymous
\else
    \section{Acknowledgments}

    This work was partially supported by the National Science and Technology Major Project under Grant No. 2023YFF0905400.
    
\fi

\bibliography{aaai2026}

\ifdefined\aaaianonymous

\newcommand{\isChecklistMainFile}{}
\input{ReproducibilityChecklist.tex}

\else
\fi

\end{document}